\documentclass[conference]{IEEEtran}
\IEEEoverridecommandlockouts
\usepackage{cite}
\usepackage{amsmath,amssymb,amsfonts}
\usepackage{algorithmic}
\usepackage{graphicx}
\usepackage{textcomp}
\usepackage{xcolor}
\def\BibTeX{{\rm B\kern-.05em{\sc i\kern-.025em b}\kern-.08em
    T\kern-.1667em\lower.7ex\hbox{E}\kern-.125emX}}

\begin{document}

\title{Randomness and Interpolation Improve Gradient Descent: A Simple Exploration in CIFAR Datasets}

\author{\IEEEauthorblockN{ Jiawen Li\IEEEauthorrefmark{1}, Pascal LEFEVRE\IEEEauthorrefmark{1}, Anwar PP Abdul Majeed\IEEEauthorrefmark{2}}
\IEEEauthorblockA{
\begin{tabular}{@{}c@{}}
\IEEEauthorrefmark{1}\textit{School of AI and Advanced Computing, XJTLU Entrepreneur College (Taicang),} \\
\textit{Xi’an Jiaotong-Liverpool University, Suzhou 21523, China}
\end{tabular}
}

\IEEEauthorrefmark{2}\textit{School of Engineering and Technology, Sunway University, Bandar Sunway, Malaysia} \\
{0009-0002-1449-2803, 0009-0001-8550-9329, 0000-0002-3094-5596}
}
\maketitle

\begin{abstract}
Based on Stochastic Gradient Descent (SGD), the paper introduces two optimizers, named Interpolational Accelerating Gradient Descent (IAGD) as well as Noise-Regularized Stochastic Gradient Descent (NRSGD). IAGD leverages second-order Newton Interpolation to expedite the convergence process during training, assuming relevancy in gradients between iterations. To avoid over-fitting, NRSGD incorporates a noise regularization technique that introduces controlled noise to the gradients during the optimization process. Comparative experiments of this research are conducted on the CIFAR-10, and CIFAR-100 datasets, benchmarking different CNNs(Convolutional Neural Networks) with IAGD and NRSGD against classical optimizers in Keras Package. Results demonstrate the potential of those two viable improvement methods in SGD, implicating the effectiveness of the advancements.
\end{abstract}

\begin{IEEEkeywords}
Deep Learning, Optimization, Stochastic Gradient Descent(SGD), Interpolation.
\end{IEEEkeywords}

\section{Introduction}
Deep learning has emerged as a dominant approach for addressing complex problems in diverse domains such as computer vision, natural language processing, and speech recognition. And it heavily relies on the optimization algorithms employed to train deep neural networks. Hence optimizing machine learning models during training is crucial for improving their performance and efficiency in various applications. SGD is the most commonly used optimization algorithm due to its simplicity and efficiency, making it a good prototype to explore how to further improve optimizers. Yet SGD has been widely used, it still holds some limitations that also could appear in other optimizers, including the potential to become trapped in suboptimal local minima and slow convergence in specific scenarios \cite{b1}, and this situation is even more apparent when SGD training with an inappropriate learning rate \cite{b11}.

Using interpolation to accelerate the process of training by assuming the gradients of each iteration are linear dependent, this sort of SGD variant is suggested as IAGD in this paper. The IAGD optimizer leverages second-order Newton interpolation to expedite the convergence process by assuming the relevance of gradients between iterations, allowing for the construction of difference quotients.

\begin{figure}
    \centering
    \includegraphics[width=0.77\linewidth]    
    {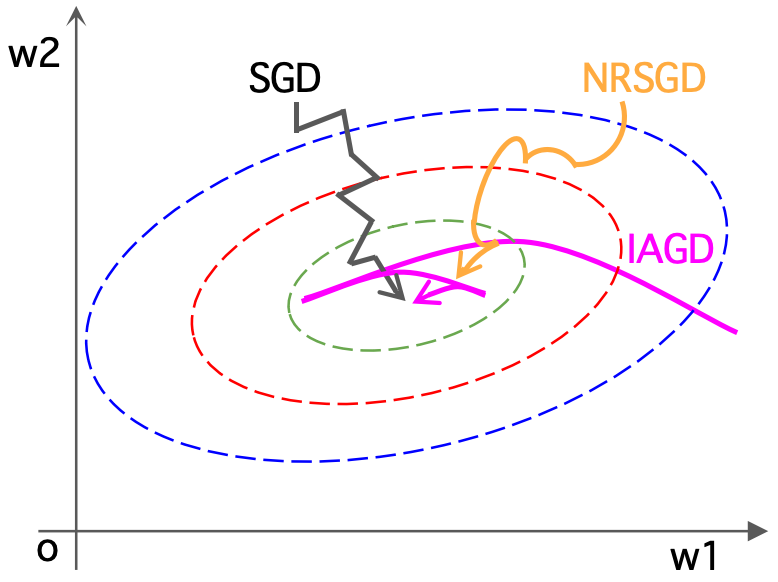}
    \caption{Except Convergence Process}
    \label{fig:1}
\end{figure}

Adding randomness is for making the optimizer get out of the local optimum in a possible convex hull. This paper proposes a novel optimization algorithm known as NRSGD. Unlike traditional first-order optimizers that rely solely on gradient information, NRSGD incorporates the second-order information by introducing random variables that follow a uniform to normal distribution instead of using momentum. These random variables are used to update the weights, with the mean and sigma values of the distribution determining the magnitudes of second-order relationships between each feature. Therefore, if the assumption is established, the convergence process for three optimizers will approximate the situation in Fig.\ref{fig:1}, in which IAGD holds the fastest convergence rate and NRSGD would show some shifting that does not direct to the optimum for exploring more potential sample space.

\section{Related Works}
The well-known Keras library provides an accessible toolkit to design as well as train deep learning models with various classical optimizers \cite{b9}. These optimization algorithms in Keras significantly affect the efficiency and effectiveness of the training process, they have the characteristics that make them applicable to a certain neural network architecture or dataset, hence it is considered a pretty reasonable setting for comparative study.

SGD is a fundamental optimization algorithm commonly used in training machine learning models. It updates the model parameters based on the gradient of the loss function computed on a subset of the training data \cite{b2}.

Adaptive Gradient Algorithm (AdaGrad) is an adaptive optimization algorithm that adjusts the learning rates of individual parameters based on the historical gradients \cite{b3}. AdaGrad performs well on sparse data and ensures that the learning rate is scaled appropriately for each parameter.

Adam optimizer combines adaptive learning rates and momentum to expedite convergence and prevent local minima, making it a popular choice for training neural networks \cite{b4}. Adagrad adjusts the learning rate for each parameter based on the historical gradients, making it suitable for sparse datasets \cite{b4}.

RMSprop uses the exponential moving average of squared gradients to adjust the learning rate, addressing the early decay issue in Adagrad \cite{b5}. Momentum optimizer introduces a momentum term to accelerate the gradient descent process and reduce fluctuations during parameter updates \cite{b6}. Those are all Keras optimizers that have been widely used and are easy to construct.

\section{Methodology}
\subsection{Noise-Regularized Stochastic Gradient Descent (NRSGD)}
NRSGD is a simple variant of the SGD optimizer that combines SGD with a parametric estimation of gradient distribution proposed to avoid over-fitting and under-fitting. It could avoiding those problems due to loss of the information with involvement of randomness or noise, this principle is somewhat similar to the use of the dropout function \cite{b13}. 

The updated formula of NRSGD is shown in Eq. \eqref{eq:1} and Eq. \eqref{eq:2}. The updating of weights not only depends on the gradient transmitted from the last layer but also relies on the random tensor $n$, adding randomness to the original SGD. This approach reducing the dependence on the gradient, makes NRSGD well suited for calculation error during the GPU acceleration as well as better optimum compared to SGD.
\begin{equation}
    \label{eq:1}
    x_{i+1}=x_{i}-\eta_{i}(w(\nabla-n)+n) ,\quad i=1,2,3...N\tag{1}
\end{equation}

\begin{equation}
    \label{eq:2}
    \eta_i = \eta_0 \times \alpha^{\left(\frac{i}{\text{s}}\right)}\tag{2}
\end{equation}

\begin{equation}
    \label{eq:3}
    n \sim N(\bar\nabla,\sigma_{\nabla})  \tag{3}
\end{equation}

To analyze the formula from another perspective, the Eq. \eqref{eq:1} could be converted as Eq. \eqref{eq:4}. In Eq. \eqref{eq:4}, the difference between gradient and noise $\Delta_{\nabla}$ is a simple approximation of differences between the gradients(Eq. \eqref{eq:5}), making the NRSDG contains second-order information.

\begin{equation}
    \label{eq:4}
    E(x_{i+1})=x_{i}-\eta(w\Delta_{\nabla}+\bar\nabla)\tag{4}
\end{equation}

\begin{equation}
    \label{eq:5}
    \Delta_{\nabla}=\nabla-n  \tag{5}
\end{equation}

Again, the inspiration of NRSGD is using a normal distribution to stimulate the real gradient as the Eq. \eqref{eq:6} shown. The NRSGD was created based on the concept that the $\nabla=n$ holds in major situations or possibilities, in that case, the NRSGD(Eq. \eqref{eq:1}) is equivalent to the SGD(in Eq. \eqref{eq:6}), which explains the inner mechanism of NRSGD.
\begin{equation}
    \label{eq:6}
    (\nabla=n) \Rightarrow (x_{i+1}=x_{i}-lr\nabla)\tag{6}
\end{equation}

Besides, the equation(1) could also explain in another aspect. Extracting the term $(1-w)$ from it, the Eq. \eqref{eq:1} converts to the Eq. \eqref{eq:7}, indicating the updates of NRSGD is actually a weighted sum of the gradient $\nabla$ and the random tensor $n$, or what we called as noise, that reducing the dependency to the gradient. This explains the robustness of NRSGD for dealing with computational errors. The reason for using Eq. \eqref{eq:1} as the NRSGD's general formula is the fewer computational times compared to Eq. \eqref{eq:7}.
\begin{equation}
    \label{eq:7}
    x_{i+1}=x_{i}-\eta_{i}(w\nabla+(1-w)n)\tag{7}
\end{equation}

\subsection{Interpolational Accelerating Gradient Descent (IAGD)}
IAGD is another variant of SGD, but combining different orders of Newton interpolation to predict the next step gradient and update it in advance, makes it theoretically speaking have a higher convergence rate to the training set, which is the main proposal for its design. This core mechanism can be expressed in the form of a function as the Eq. \eqref{eq:8} shown. To be precise, the next gradient depending the last gradient as one of the parameters.

\begin{equation}
    \label{eq:8}
    \bigtriangledown_{t+1}=g(\bigtriangledown_t),\quad t=1,2,3...T\tag{8}
\end{equation}

Based on this assumption, the majority of optimizers could all use interpolation to approximate the next step gradient in advance and use this predicted gradient in the updates of the weights. The research will utilize the Newton Interpolation(seen in general form in Eq. \eqref{eq:9} with order $r$) in the SGD, which is IAGD.
\begin{equation}
    \label{eq:9}
    N_r(x) = \sum_{i=0}^{r} \left[ f[x_0, x_1, ..., x_i] \cdot \prod_{j=0}^{i-1} (x - x_j) \right] ,\quad r \in Z^+
    \tag{9}
\end{equation}

\begin{equation}
    \label{eq:10}
    N_r(x)\approx g(\bigtriangledown_t)
    \tag{10}
\end{equation}

Yet the interpolation provides a precise acceleration during the optimization, the cost of the computing cannot be ignored when encountering large datasets, hence the IAGD adopts second-order Newton Interpolation as the equation \eqref{eq:11} shown.

\begin{equation}
    \label{eq:11}
    N_2(x)=x_0+f[x_0,x_1,x_2]*(x-x_0)+f[x_1,x_2]*(x-x_0)*(x-x_1)
    \tag{11}
\end{equation}

By merging those conditions, the general formulas of the IAGD are constructed. The expression compose of three terms with different quotients, first term shows the initial gradient with the coefficient 1, the second term $P_1$ \eqref{eq:13} has the difference quotients between $\bigtriangledown_{t-4}$ and $\bigtriangledown_{t-3}$, and the $P_2$ is term with second order quotients as the Eq. \eqref{eq:14} explicate. Eq. \eqref{eq:15} to \eqref{eq:17} shows the decompose terms of P1 and P2.  

\begin{equation}
    \label{eq:12}
    N_2(\bigtriangledown_l)=\bigtriangledown_{l-4} + P_1 +P_2
    \tag{12}
\end{equation}
\begin{equation}
    \label{eq:13}
    P_1=g[\bigtriangledown_{t-4},\bigtriangledown_{t-3}]
    (\bigtriangledown_l-\bigtriangledown_{t-4})
    \tag{13}
\end{equation}

\begin{equation}
    \label{eq:14}
    P_2=g[\bigtriangledown_{t-4},\bigtriangledown_{t-3},\bigtriangledown_{t-2}](\bigtriangledown_t-\bigtriangledown_{t-4})(\bigtriangledown_t-\bigtriangledown_{t-3})
    \tag{14}
\end{equation}

\begin{equation}
    \label{eq:15}
    g[\bigtriangledown_{t-4}, \bigtriangledown_{t-3}]=\frac{g(\bigtriangledown_{t-3})-g(\bigtriangledown_{t-4})}{\bigtriangledown_{t-3}-\bigtriangledown_{t-4}}=\frac{\bigtriangledown_{t-2}-\bigtriangledown_{t-3}}{\bigtriangledown_{t-3}-\bigtriangledown_{t-4}}
    \tag{15}
\end{equation}

\begin{equation}
    \label{eq:16}
    g[\bigtriangledown_{t-3}, \bigtriangledown_{t-2}]=
    \frac{\bigtriangledown_{t-1}-\bigtriangledown_{t-2}}{\bigtriangledown_{t-2}-\bigtriangledown_{t-3}}
    \tag{16}
\end{equation}

\begin{equation}
    \label{eq:17}
    g[\bigtriangledown_{t-4}, \bigtriangledown_{t-3}, \bigtriangledown_{t-2}]= \frac{g[\bigtriangledown_{t-3}, \bigtriangledown_{t-2}] - g[\bigtriangledown_{t-4}, \bigtriangledown_{t-3}]}{\bigtriangledown_{t-2} - \bigtriangledown_{t-4}}
    \tag{17}
\end{equation}

The IAGD's expression for updates could simplified as the form of Eq. \eqref{eq:20} as well as Eq. \eqref{eq:21}. To be precise, the identity of the IAGD is using the interpolation to approximate the gradient of the next step and update it in advance to accelerate the training process, the Eq. \eqref{eq:18} illustrates the ideal situation of IAGD, making one step update have the same effect as two.
\begin{equation}
    \label{eq:18}
     x_{i+1}=x_{i}-\eta_{i}\bigtriangledown_{i}-\eta_{i-1}\bigtriangledown_{i+1},\quad i=2,4,6...N
    \tag{18}
\end{equation}

\begin{equation}
    \label{eq:19}
    \bigtriangledown_{i+1}\approx f(\bigtriangledown_{i})
    \tag{19}
\end{equation}

Besides, due to the assumption in Eq. \eqref{eq:19}, this interpolation needs four pieces of gradients to estimate. For reducing the error between the real values and the interpolations, the IAGD uses one-order Newton Interpolation to approximate when the information is not enough(when i=3), which means there's no term $P_2$ in the Eq. \eqref{eq:21}. As for the updates of the learning rate, the IAGD also uses the default decay methods that the package Tensorflow provides, the Eq. \eqref{eq:1} shows it in detail.
\begin{equation}
    \label{eq:20}
    x_{i+1}=x_{i}-(\eta_{i}\bigtriangledown_{i}+\eta_{i-1}f(\bigtriangledown_{i})),\quad i=2,4,6...N
    \tag{20}
\end{equation}

\begin{equation}
    \label{eq:21}
    f(\bigtriangledown_{i}) =
    \left\{
    \begin{aligned}
    \bigtriangledown_{i} \quad (when \quad i<3)
    \\
    (\bigtriangledown_{i-3} + P_1) \quad (when \quad i=3)
    \\
    (\bigtriangledown_{i-4} + P_1 +P_2) \quad (when \quad i>3)
     \end{aligned}
    \tag{21}
    \right.
\end{equation}

\section{Experiment}
\subsection{Experimental Design}
The study selects two famous datasets for testing NRSGD and IAGD, CIFAR-10, and CIFAR-100. CIFAR-10 is a widely used image classification dataset that consists of 10 different classes. Each class contains 6000 32x32 color images. CIFAR-100 is a more challenging image classification dataset that includes 100 different classes, each class contains 600 32x32 color images \cite{b7}.

Since all CIFAR datasets, datasets were selected based on their similarities and increasing difficulty levels. Restricted by the device, the comparative analysis involved assessing the effectiveness of IAGD in comparison to traditional optimizers such as Adam, SGD, and RMSprop within the same convolutional network architecture including AlexNet and LeNet5.

\begin{figure}[h]
    \centering
    \includegraphics[width=0.78\linewidth]
    {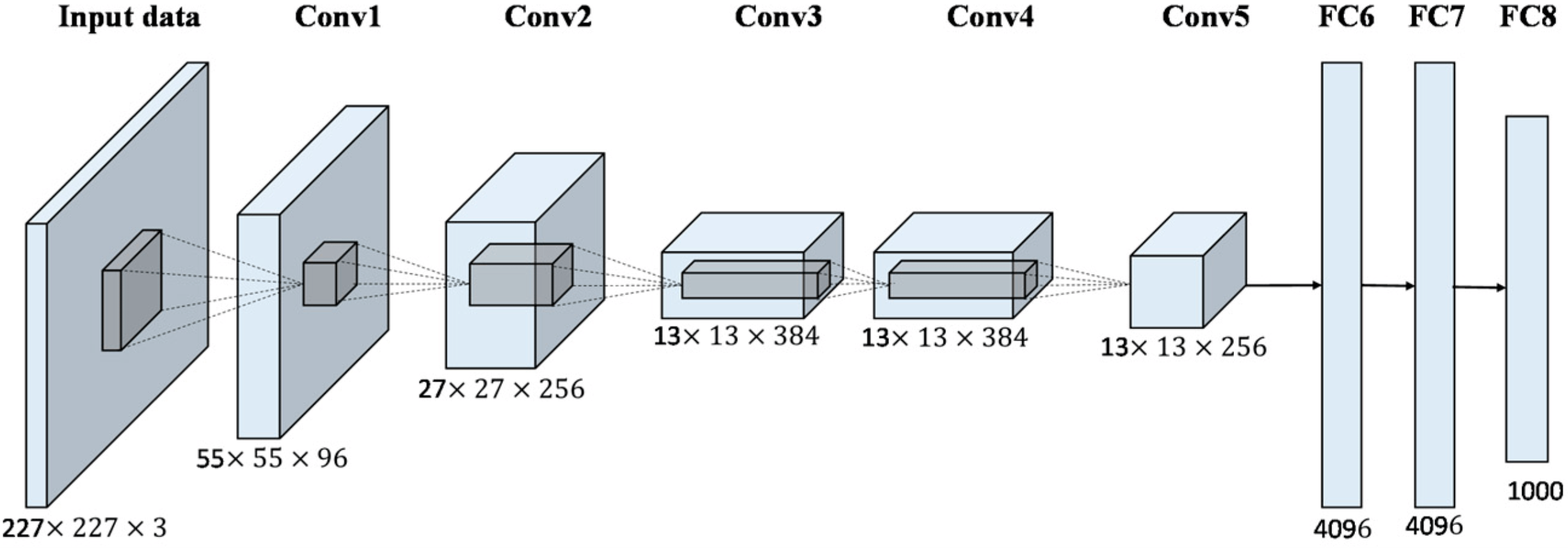}
    \caption{AlexNet Framework \cite{b8}}
    \label{fig:2}
\end{figure}

\begin{figure}[h]
    \centering
    \includegraphics[width=0.78\linewidth]
    {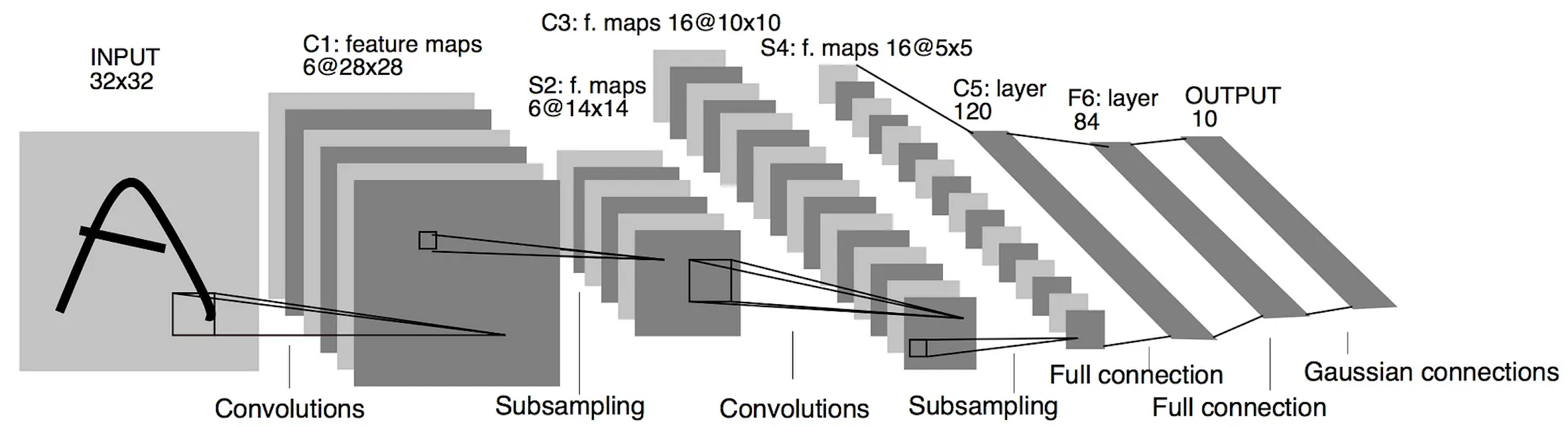}
    \caption{LeNet-5 Framework \cite{b10}}
    \label{fig:3}
\end{figure}

As Fig.\ref{fig:2} shown, AlexNet is a famous convolutional neural network architecture that reached state of the art in image classification task \cite{b12}. Comprising of multiple convolutional layers, a pooling layer, and a fully connected layer. LeNet5 is also a famous and relatively simple convolutional neural network architecture that is good for some image classification problems \cite{b10}(seen details in Fig.\ref{fig:3}). Our motivation for choosing those architectures is due to the models have gained prominence throughout image classification. These architectures have been studied in a myriad of works and are relatively robust, making them reasonable to test optimizers.

The experiments involved training the models for 200 epochs with a consistent initial learning rate($\eta_0$) of 0.001 and using cross-entropy as the loss function across all optimizers, each image would be resized to shape (112,112,3), allowing the data could be processed by AlexNet. The datasets were partitioned into a 1:9 ratio for validation and training within the training set, and a 1:5 ratio for the test set. For replicating the experiment, the paper provides a complete code of it \cite{code}.

\begin{equation}
    \label{eq:22}
    L = -\frac{1}{N} \sum_{i=1}^{N} y_i \log(p_i) + (1-y_i) \log(1-p_i)
    \tag{22}
\end{equation}

\subsection{Performance Analysis}
\begin{figure}[h]
    \centering
    \includegraphics[width=0.85\linewidth]
    {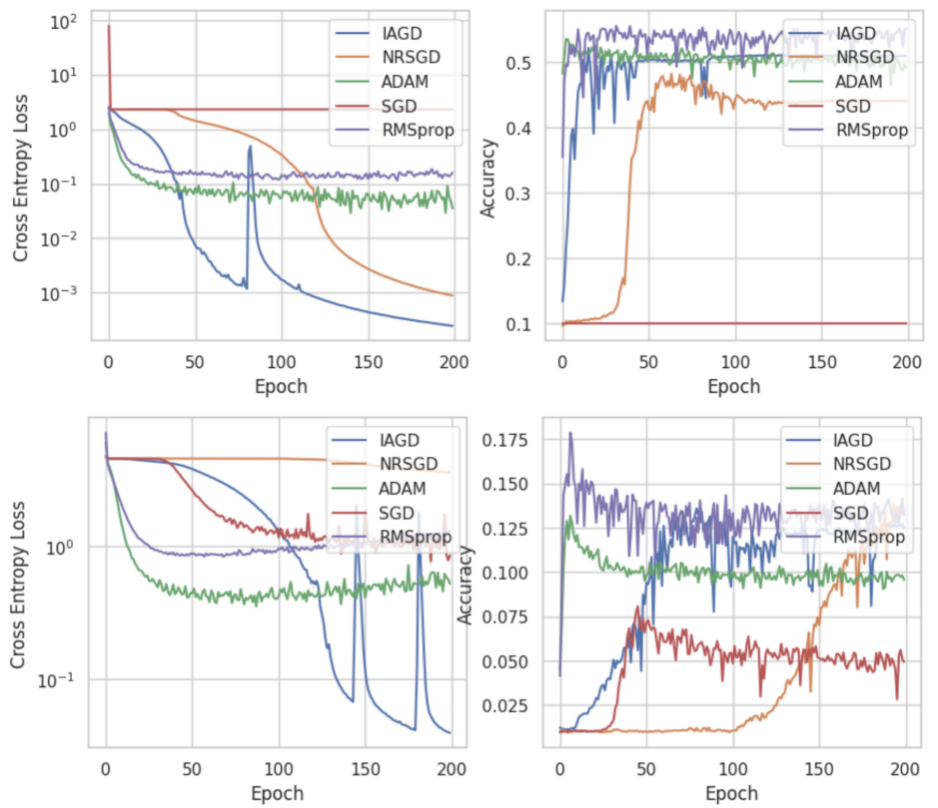}
    \caption{Performance of LeNet-5 with Tested Optimizers}
    \label{fig:LeNet-CIFAR}
\end{figure}

\begin{figure}[h]
    \centering
    \includegraphics[width=0.85\linewidth]
    {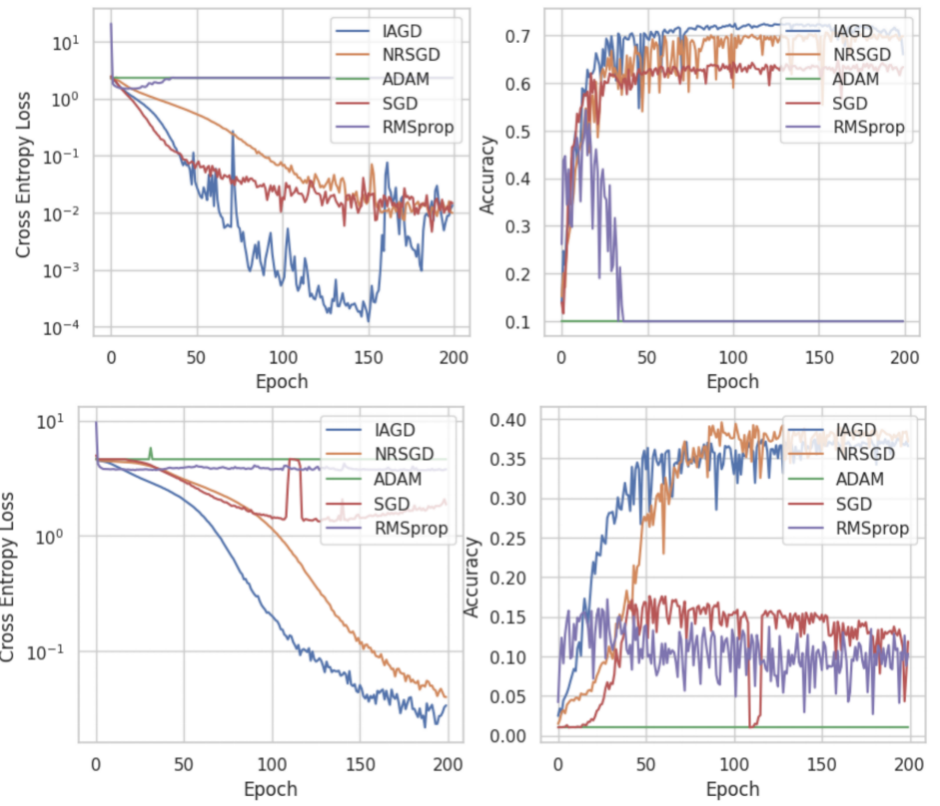}
    \caption{Performance of AlexNet with Tested Optimizers}
    \label{fig:Alex-CIFAR}
\end{figure}

For Fig.\ref{fig:LeNet-CIFAR} and Fig.\ref{fig:Alex-CIFAR}, the subplot on the left column represents the cross-entropy loss(seen in Eq. \eqref{eq:22}) of the train set, and the right column shows the accuracy of the validation set. Additionally, the first row of the plots is tested in the CIFAR-10 dataset, and the second row is for CIFAR-100.

Fit with the previous assumptions, the IAGD holds the smallest loss during the training for all tested situations. But when optimizing with LeNet-5, IAGD does not converge fast in the initial, which is possibly caused by the lack of information to calculate the divided difference, making it perform approximate SGD in the very beginning. Also, IAGD shows a large fluctuation, which might indicate a simple early stopping could improve it further.

\begin{table}[h]
\caption{The Performance of LeNet-5 in Test Set}
\begin{center}
\setlength{\tabcolsep}{2pt}
\begin{tabular}{|c|ccc|ccc|}
\hline
\multicolumn{1}{|c|}{{\textbf{Optimizers}}} & \multicolumn{3}{c|}{\textbf{CIFAR-10}} & \multicolumn{3}{c|}{\textbf{CIFAR-100}} \\
\cline{2-7}
& Accuracy& Loss$^{\mathrm{a}}$& Time(s)$^{\mathrm{b}}$& Accuracy & Loss& Time(s)\\
\hline
$IAGD$&0.5097&7.4239 &1305.77 
&0.1248 &19.0587 &\textbf{1271.01}$^{\mathrm{c}}$
\\
$NRSGD$&0.4408&10.2424 &1368.75
&\textbf{0.1326} &\textbf{3.9328} &1349.15
\\
$ADAM$&0.4917&10.1348 &1320.79 
&0.0956 &110.7334 &1404.50
\\
$SGD$&0.1000&\textbf{2.3025}&\textbf{1293.69} 
&0.0493 &22.8118 &1388.81
\\
$RMSprop$&\textbf{0.5516}&30.0231 &1306.50 
&0.1321 &167.1602 &1309.63
\\
\hline
\multicolumn{4}{l}{$^{\mathrm{a}}$The Cross-Entropy Loss}
\\
\multicolumn{5}{l}{$^{\mathrm{b}}$The time cost during training, the unit of it is second}
\\
\multicolumn{6}{l}{$^{\mathrm{c}}$The bolded data are the smallest loss/time or the largest accuracy}
\end{tabular}

\label{tab1}
\end{center}
\end{table}

\begin{table}[h]
\caption{The Performance of AlexNet in Test Set}
\setlength{\tabcolsep}{2pt}
\begin{center}
\begin{tabular}{|c|ccc|ccc|}
\hline
\multicolumn{1}{|c|}{{\textbf{Optimizers}}} & \multicolumn{3}{c|}{\textbf{CIFAR-10}} & \multicolumn{3}{c|}{\textbf{CIFAR-100}} \\
\cline{2-7}
& Accuracy& Loss&Time(s)& Accuracy & Loss&Time(s)\\
\hline
$IAGD$&0.6602 &3.1540 &\textbf{2466.93}
&0.3656 &6.8141 &\textbf{2518.09}
\\
$NRSGD$&\textbf{0.6975} &2.9031 & 3158.57
&\textbf{0.3713} &6.6966 &3200.64
\\
$ADAM$&0.1000 &\textbf{2.3026} & 2736.92
&0.0100 &4.6051 &2811.24
\\
$SGD$&0.6335 &2.8431 & 2551.53
&0.1187 &4.3746 &2599.46
\\
$RMSprop$&0.1000 &\textbf{2.3026} & 2549.90
&0.1033 &\textbf{3.7843} &2609.68
\\
\hline
\end{tabular}
\label{tab2}
\end{center}
\end{table}

CIFAR datasets are all balanced datasets, which means they have an equal number of class samples, making accuracy metrics extremely valuable in evaluation. Except for the case in CIFAR-10, as TABLE \ref{tab1} shown(accuracy:0.4408), the models with NRSGD have the highest accuracy in the test, indicating its superior performance. The IAGD optimizer presents as the fastest in tested situations.

As a result, the IAGD and NRSGD proposed in this study can improve the comprehensive performance of the CNN models in tested CIFAR datasets compared to SGD. Specifically, the IAGD holds the shortest training time with good accuracy, and NRSGD holds the best performance among the tested optimizers. Both of them are stable and do not under-fitting as the others do, like ADAM having this problem except when optimized LeNet-5 with CIFAR-10 dataset, SGD explicates its effectiveness with AlexNet in CIFAR-10 task(seen in TABLE \ref{tab2}), and RMSProp only works when optimized LeNet.

\section{Conclusion}
In summary, the paper introduces IAGD and NRSGD, presenting unneglectable enhancements in the experiment. This potential improvement in the convergence process and exploration not only illustrated the merits of suggested optimizers but could also have a significant implication for improving other optimizers. Future work could focus on further validation of these enhancements across diverse datasets and architectures, exploring their applicability to large-scale tasks.

\end{document}